# Segmentation by Test-Time Optimization (TTO) for CBCT-based Adaptive Radiation Therapy

Xiao Liang[1], Jaehee Chun[2], Howard Morgan[1], Ti Bai[1], Dan Nguyen[1], Justin C. Park[1], Steve Jiang[1], *

[1]Medical Artificial Intelligence and Automation Laboratory and Department of Radiation Oncology, University of Texas Southwestern Medical Center, Dallas, TX, USA

[2]Department of Radiation Oncology, Yonsei Cancer Center, Yonsei University College of Medicine, Seoul, South Korea

*E-mail: Steve.Jiang@UTSouthwestern.edu

## Highlights

- A TTO method is proposed to refine a pre-trained deformable image registration model for each individual test patient and then progressively for each fraction of online adaptive radiation therapy treatment to mitigate the model generalizability problem.
- Extensive experiments are performed to show that TTO can significantly improve a population model's performance especially for some patients where the population model fails.
- TTO methods can be accomplished in a few minutes, making it suitable for online clinical applications.

## Abstract

Online adaptive radiotherapy (ART) requires accurate and efficient auto-segmentation of target volumes and organs-at-risk (OARs) in mostly cone-beam computed tomography (CBCT) images, which often have severe artifacts and lack soft tissue contrast, making direct segmentation very challenging. Propagating expert-drawn contours from the pre-treatment planning CT (pCT) through traditional or deep learning (DL) based deformable image registration (DIR) can achieve improved results in many situations. Typical DL-based DIR models are population based, that is, trained with a dataset for a population of patients, so they may be affected by the generalizability problem. In this paper, we propose a method called test-time optimization (TTO) to refine a pre-trained DL-based DIR population model, first for each individual test patient, and then progressively for each fraction of online ART treatment. Our proposed method is less susceptible to the generalizability problem, and thus can improve overall performance of different DL-based DIR models by improving model accuracy, especially for outliers. Our experiments used data from 239 patients with head and neck squamous cell carcinoma to test the proposed method. Firstly, we trained a population model with 200 patients, and then applied TTO to the remaining 39 test patients by refining the trained population model to obtain 39 individualized models. We compared each of the individualized models with the population model in terms of segmentation accuracy. The average improvement of Dice Similarity Coefficient (DSC) and 95% Hausdorff Distance (HD95) of the segmentation can be up to 0.04 (5%) and 0.98 mm (25%), respectively, with the individualized models compared to the population model over 17 selected OARs and a target of 39 patients. While the average improvement may seem mild, we found that the improvement for outlier patients is significant. The number of patients with at least 0.05 DSC improvement or 2 mm HD95 improvement by TTO averaged over the 17 selected structures for the state-of-the-art architecture Voxelmorph is 10 out of 39 test patients. We also evaluated the efficiency gain of deriving the individualized models from the pre-trained population model versus from an un-trained model. The average time for deriving the individualized model using TTO from the pre-trained population model is approximately four minutes, which is about 150 times faster than that required to derive the individualized model from an un-trained model. We also generated the adapted fractional models for each of the 39 test patients by progressively refining the individualized models using TTO to CBCT images acquired at the later fractions of online ART treatment. When adapting the individualized model to a later fraction of the same patient, the average time is reduced to about one minute and the accuracy is slightly improved. The proposed TTO method is well-suited for online ART and can boost segmentation accuracy for DL-based DIR models, especially for outlier patients where the pre-trained models fail.

## Graphical abstract

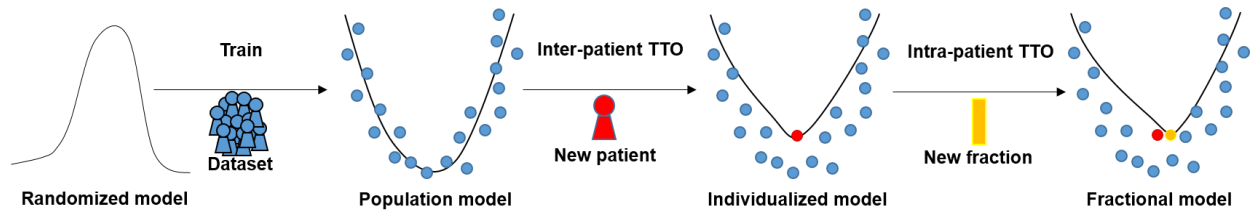

## 1. Introduction

Online adaptive radiotherapy (ART) is an advanced radiotherapy technology in which the daily treatment plan is adapted to the patient's changing anatomy (e.g., shrinking tumor, fluctuations in body weight), typically using cone beam computed tomography (CBCT) images. The online nature of the treatment demands high efficiency since the patient is immobilized to the treatment position while waiting for treatment to start. The time-consuming process of segmenting the tumor volumes and organs at risk (OARs) has become a major bottleneck for the widespread use of online ART, warranting an urgent need for accurate auto-segmentation tools (Glide-Hurst et al., 2021).

Auto-segmentation of CBCT images is a very challenging task, mainly due to severe artifacts including low soft-tissue contrast and image truncations (Glide-Hurst et al., 2021). Currently, there are two main categories of CBCT auto-segmentation methods for online ART: deformable image registration (DIR) based and deep learning (DL) based (Cardenas et al., 2019). The DIR-based auto-segmentation is widely used in the ART clinical workflow (Glide-Hurst et al., 2021; Veresezan et al., 2017; Zhang et al., 2007). It deforms the pre-treatment planning CT (pCT) image which has the target and OAR contours pre- determined by experts, to the CBCT image, which is based on the treatment plan that is adapted to the new anatomy. The resulted deformation vector field (DVF) is then used to propagate the contours from pCT to CBCT. Evaluation of different DIR algorithms for contour propagation between pCT and CBCT in head & neck (H&N) ART suggests that careful examinations and modifications are still required (Li et al., 2017). DL-based auto-segmentation has achieved clinically acceptable performance in many image modalities (Cardenas et al., 2019), such as CT, while DL-based direct segmentation in CBCT images is still very challenging due to poor image quality. Consequentially, a hybrid auto-segmentation approach for CBCT-based online ART has been implemented in clinical practice (Sibolt et al., 2021). This hybrid approach uses a DL-based model to direct segmentation to easier OARs in CBCT images, and subsequently uses the segmentation results to constrain the DIR between pCT and CBCT, propagating the target and rest of OARs from pCT and CBCT. Although this method works for some OARs and targets, manual editing of challenging OARs and target volumes remains required and time-consuming.

Recently, DL-based DIR methods have shown excellent performance in many applications over popular traditional DIR methods including B-spline algorithms, represented by ELASTIX (Klein et al., 2010) and 3DSlicer B-spline registration (Fedorov et al., 2012), as well as Demons algorithms (Gu et al., 2009; Vercauteren et al., 2009). In 2015 Jaderberg *et al* proposed a spatial transformer network (STN) that is differentiable and allows for spatial transformations on the input image inside a neural network, while having the ability to be added to any other existing architecture (Jaderberg et al., 2015). As a result, the STN network has inspired many unsupervised DIR methods. A typical unsupervised DIR model can be divided into two parts: DVF prediction and spatial transformation. In DVF prediction, a neural network takes a pair of fixed and moving images as input and then outputs a DVF. Subsequently in spatial transformation, the STN network warps the moving image according to the predicted DVF to achieve the moved image. The loss function for model training is usually composed of image similarity loss between the fixed and moved images as well as a regularization term on DVF. Voxelmorph, proposed by Dalca *et al,* combined a probabilistic generative model and a DL model for diffeomorphic registration (Dalca et al., 2019). They used a U-Net architecture to predict velocity field and diffeomorphic integration layers to sample DVF from the predicted velocity field, followed by a STN network to warp the moving images. Following this, image similarity and Kullback-Leibler divergence constraints were used in the loss function. A similar work, FAIM, used a U-Net architecture to predict DVF directly and a STN network to warp images (Kuang and Schmah, 2019). The loss function of FAIM is also composed of image similarity and regularization terms to constrain DVF smoothness. To further improve the performance of unsupervised DL methods, Zhao *et al* built recursive cascaded networks (Zhao et al., 2019) on top of a base network including VTN (Zhao et al., 2020) and Voxelmorph (Balakrishnan et al., 2018). The cascade procedure is done by recursively performing registration on warped images, and the final DVF is a composition of all predicted DVFs. Importantly, these results showed that recursive cascaded networks outperform the base network with significant gains.

These DL models for DIR are all population-based, that is, they are trained on a dataset representing a population of patients. Generalizability problems may exist in these models when deployed to patients where the distribution of inputs differs from that of the training dataset. In the targeted clinical applications of this work, many factors could cause such a problem, including different anatomical sites, scanning machines, and scanning protocols. Therefore, the model's generalizability problem needs to be carefully and thoroughly addressed.

To solve this problem, we propose a method called test-time optimization (TTO) to individualize a pre-trained population DL model for one pair of fixed and moving images, by iteratively refining the weights of the DL model in a traditional optimization matter. TTO was inspired by the work of both Chen *et al* (Chen et al., 2020) and Fechter *et al* (Fechter and Baltas, 2020), where one-shot learning is used for DIR to generate anthropomorphic phantoms and to track periodic motion with DL models, respectively. The predicted DVF is then used to warp the moving image to match the fixed image. Essentially, TTO overfits the DL model to a specific pair of moving and fixed images, avoiding generalization problems and promising better performance compared to the direct use of the population DL model. When TTO is applied to a pre-trained DL model versus applied to an untrained model as in the work of Chen *et al* (Chen et al., 2020) and Fechter *et al* (Fechter and Baltas, 2020), improved efficiency is also expected, which is critical for online applications. For CBCT-based online ART, TTO can be used to refine a pre-trained DL model to a new patient (*inter-patient TTO*) to get an individualized model and also to further refine the individualized model to a new treatment fraction for the same patient (*intra-patient TTO*) to get a fractional model.

In the following content, we first introduce the common architecture used in unsupervised DL-based DIR algorithms in Section 2.1. Following that, we introduce the concept of inter- and intra- patient TTO methods that can be applied to the unsupervised DIR algorithms in Section 2.2. Lastly, we describe the data used in the experiments and the experiment design in Section 2.3 and Section 2.4. Three main experiments have been performed in this study: 1) We compared the performance of the individualized TTO model with the population model for different DL architectures 2) We compared the efficiency of inter-patient TTO starting from a pre-trained population model and an untrained model with random weights for the two best DL architectures 3) We further refined the individualized model to a later treatment fraction of the same patient to obtain a fractional model to illustrate intra-patient TTO application in CBCT-based online ART workflow. We present the results of the three experiments in Section 3, and the conclusions and discussion in Section 4.

Our main contributions are:

1. We proposed a TTO method that can refine a pre-trained DL-based population DIR model for each individual test patient and then progressively for each fraction of online ART treatment to mitigate the model generalizability problem.

2. We performed extensive experiments for multiple state-of-the-art DL architectures to show that TTO can significantly improve a population model's performance, especially when the population model doesn't work well for a particular patient.
3. We showed that TTO models are less susceptible to the generalizability problem, a common issue for the population models in the targeted clinical applications.
4. We demonstrated that the individualized models from TTO outperform the population DL models and traditional DIR models.
5. We showed that both the inter- and intra-patient TTO can be accomplished in a few minutes.
6. We also showed that inter- and intra-patient TTO can be applied to DIR in an online ART workflow for auto-segmentation effectively and efficiently by adapting a population model to a new patient or adapting an individualized model to a new treatment fraction of the same patient.

## 2. Materials and methods

### 2.1. Unsupervised DL-based DIR algorithms

Two pairs of images are set to be the moving images $I_m$ and the fixed images $I_f$; we assume that they are pre-rigid aligned. DIR tries to find the best DVF $u$ that can minimize the difference between the fixed and moved images. Thus, an ideal DIR between $I_m$ and $I_f$ can be expressed as:

$$I_f(x) = I_m \circ u,$$

where $I_m \circ u$ denotes $I_m$ warped by $u$.

The typical unsupervised DL-based DIR algorithm is shown in Figure 1. A transformation neural network is used to predict DVF from a pair of moving and fixed images, followed by use of a STN to warp the moving images based on a predicted DVF to obtain deformed moving images. The transformation model can be any neural network, either a very simple one like a 10-layer convolutional neural network (CNN), or a state-of-the-art architecture like Voxelmorph (Balakrishnan et al., 2018; Dalca et al., 2019) or cascaded VTN (Zhao et al., 2019). The loss function can be described as:

$$\mathcal{L} = \mathcal{L}_{sim}(I_f, I_m \circ u) + \lambda R(u) = \mathcal{L}_{sim}\left(I_f, I_m \circ f(I_m, I_f)\right) + \lambda R\left(f(I_m, I_f)\right),$$

where $\mathcal{L}_{sim}$ is the image similarity measure between fixed and deformed moving images, $R$ is the regularization term for $u$, and $\lambda$ is a weighting factor. In our study, $I_m$ is pCT and $I_f$ is CBCT, with the DL model optimized by minimizing loss function. A population model is trained on a large dataset and after training, during the inference phase, DVF can be predicted by the population model from a pair of pCT and CBCT images. STN can then be used to warp the contours on pCT to achieve auto-segmentation on CBCT with the predicted DVF.

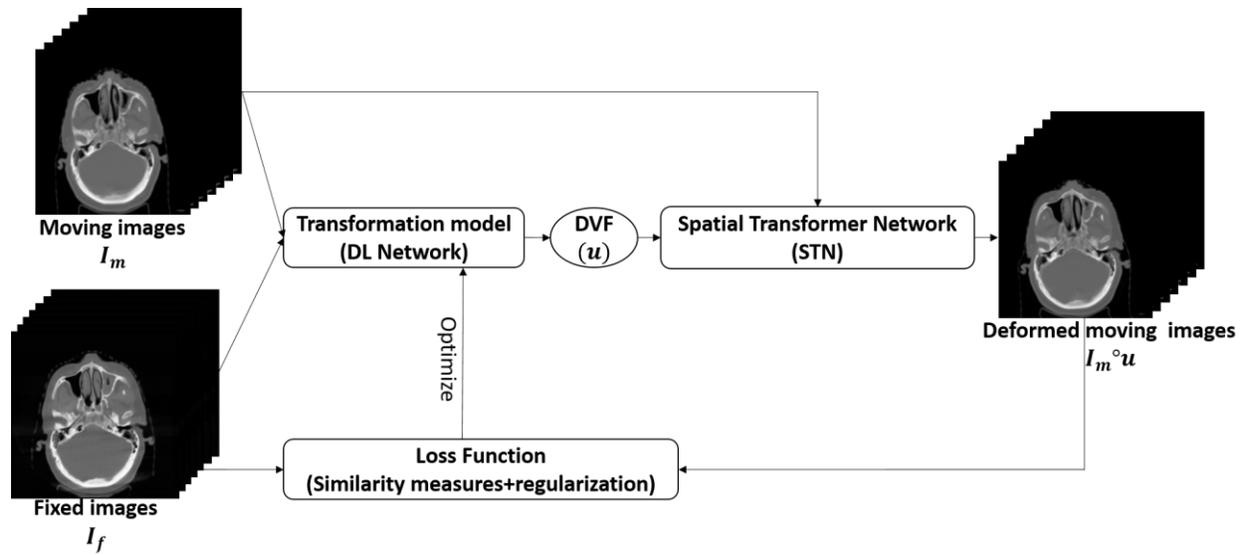

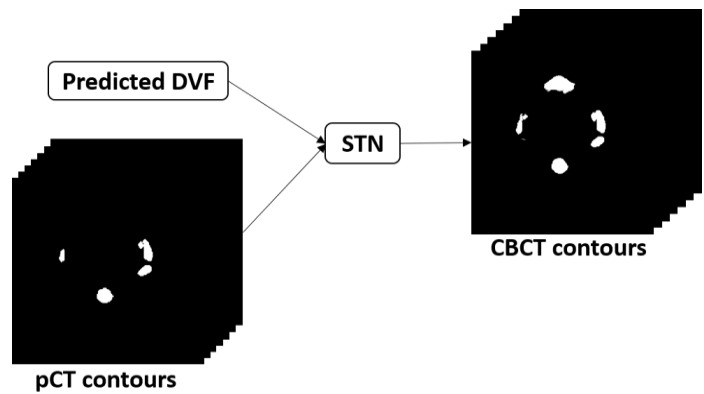

**Figure 1.** Step 1 shows a classical training architecture of unsupervised DL-based DIR algorithms. During training, the transformation model is optimized by minimizing the loss function. Step 2 shows the inference phase, where predicted DVF can be used to warp pTV contours through STN to obtain contours on CBCT.

## 2.2. Inter-patient TTO and intra-patient TTO

A feedforward network with a single layer is sufficient to represent any function, but the layer may be unfeasibly large and fail to learn and generalize correctly (Goodfellow et al., 2016). In the mathematical theory of artificial neural networks, universal approximation theorems are results (Hornik et al., 1989) that establish the density of an algorithmically generated class of functions within a given function space of interest. There are a variety of results between non-Euclidean spaces and other commonly used architectures and, more generally, algorithmically generated sets of functions, such as the CNN architecture (Heinecke et al., 2020; Zhou, 2020), radial-basis-function networks (Park and Sandberg, 1991), or invariant/equivariant network (Yarotsky, 2021). Universal approximation theorems imply that neural networks can approximate any function with appropriate capacities. Thus, according to universal approximation theorems, a DL neural network can approximate a transformation function in DIR with only a moving and a fixed image.

TTO method is to optimize a model during inference stage rather than training stage. Unlike common DL training strategies in which a DL model is trained on a large dataset to generate a population model and is then tested on an unseen dataset, TTO doesn't need pre-training on much data. Instead, only one pair of moving and fixed images is enough for a DL network to generate a transformation function for that image pair according to universal approximation theorems. In our application, the biggest advantage of TTO is that the patient-specific transformation model can be generated for each patient, rather than just one population model which is then applied for all patients. Therefore, the common generalizability problem or overfitting problem in machine learning can be avoided by the TTO strategy. However, like all optimization methods, TTO may also suffer from computation time costs. This issue can be greatly mitigated by starting TTO from a pre-trained population model rather than starting from scratch to reduce the number of iterations needed. Many studies have shown that the amount of time or iterations needed to learn an accurate neural network model can be significantly reduced by transfer learning over learning from scratch (Mihalkova et al., 2007; Shao et al., 2019; Tan et al., 2018). Therefore, the model parameters from a population model that has been trained on large datasets can be utilized in TTO to start the optimization process.

Figure 2 illustrates the concept of inter- and intra- patient TTO in our application. The first step is to obtain a population model by training a DL network on a large dataset. If a new patient's anatomy is very different from the training data, the population model might fail. However, TTO can adjust the population model parameters to the new patient to obtain the individualized model. This means we can apply TTO to a DL model starting from the population model parameters rather than starting from scratch on a new patient's data in order to achieve the best performance for that specific patient. Therefore, TTO will not only improve DIR accuracy compared to the population model, it will also decrease the number of iterations and optimization time by starting TTO from a warm start. The second step is to get an individualized model by fine-tuning the population model to a new patient.

In CBCT-based ART workflow, CBCT images are frequently taken during radiation courses for treatment setip verification or anatomical changes monitoring. This is with the assumption that a new patient has a CBCT image for each treatment fraction and that it already has an individualized model refined to the CBCT image from the first fraction. In this case, TTO can be applied to the individualized model on the image pair of the next fraction to obtain the fractional model, and so forth. Therefore, each fraction will have a fractional model that has the best fit for that fraction by TTO. The third step is to get a fractional model by fine-tuning the individualized model to a new treatment fraction of the same patient.

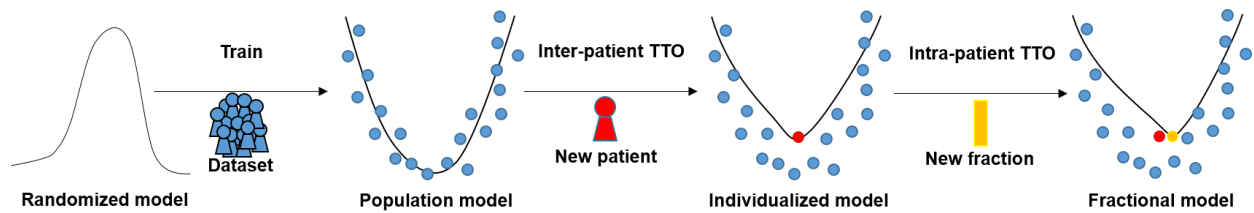

**Figure 2. Concept of the population model, individualized model (inter-patient TTO), and fraction model (intra-patient TTO).** The population model is obtained by a typical training strategy, which trains a DL model on a large dataset. Inter-patient TTO: an individualized model for a new patient can be obtained by adapting the population model to the new patient's data. Intra-patient TTO: a fractional model for the same patient can be obtained by adapting the individualized model to the new fraction's data.

## 2.3. Data

We retrospectively collected data from 239 patients with head and neck squamous cell carcinoma treated with conventionally fractionated external beam radiotherapy to a total dose of approximately 70Gy. Each patient's data included a 3D pCT volume acquired before the treatment course, OARs and target segmentations delineated by physicians on the pCT, and two sets of 3D CBCT volume acquired at fraction 20 and fraction 21 during the treatment course. The pCT volumes were acquired by a Philips CT scanner with $1.17 \times 1.17 \times 3.00$ mm$^3$ voxel spacing. The CBCT volumes were acquired by Varian On-Board Imagers with $0.51 \times 0.51 \times 1.99$ mm$^3$ voxel spacing and $512 \times 512 \times 93$ dimensions. The pCT is rigid -registered to CBCT through Velocity (Varian Inc., Palo Alto, USA), therefore the rigid-registered pCT has the same voxel spacing and dimensions as CBCT. Synthetic CT (sCT) images with less artifacts and CT-like Hounsfield units were generated from CBCT using an in-house DL model developed previously(Liang et al., 2019). In this paper, sCT replaced CBCT in all the following DIR experiments since better image quality would lead to more accurate DIR. The dimensions of pCT and sCT volumes were both down-sampled to $256 \times 256 \times 93$ from $512 \times 512 \times 93$, and then cropped to $224 \times 224 \times 64$. We randomly picked 39 out of 239 patients for testing, and selected 17 structures that were either critical OARs or had large anatomical changes during radiotherapy courses. These structures were: left brachial plexus (L_BP), right brachial plexus (R_BP), brainstem, oral cavity, constrictor, esophagus, nodal gross tumor volume (nGTV), larynx, mandible, left masseter (L_Masseter), right masseter (R_Masseter), posterior arytenoid-cricoid space (PACS), left parotid gland (L_PG), right parotid gland (R_PG), left submandibular gland (L_SMG), right submandibular gland (R_SMG), and spinal cord. The contours of these 17 structures were first propagated from pCT to CBCT of fraction 21 using rigid and deformable image registration in Velocity, and then modified and approved by a radiation oncology expert as ground truth contours on CBCT of fraction 21 for testing.

### 2.4. Experiment design

We performed extensive experimentation to answer the following: 1) Does the TTO method have any advantage over the typical training strategy? 2) How do you use TTO in an efficient way for CBCT-based online ART? The experiment design is shown in Figure 3.

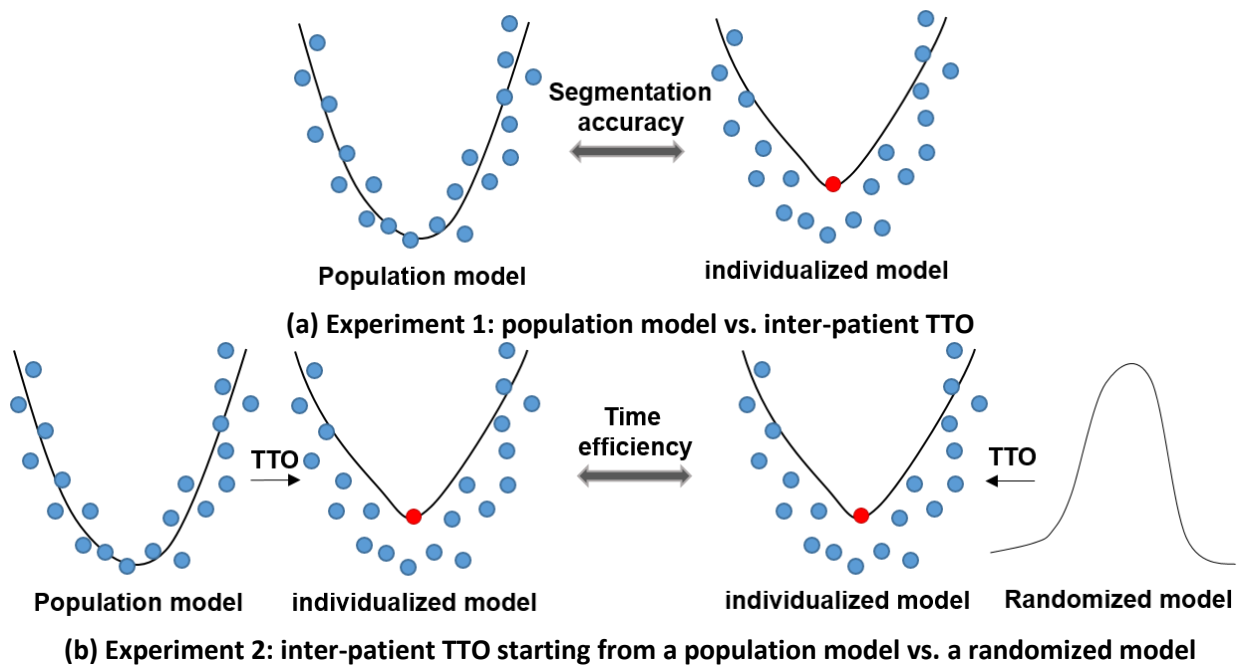

(a) Experiment 1: population model vs. inter-patient TTO

(b) Experiment 2: inter-patient TTO starting from a population model vs. a randomized model

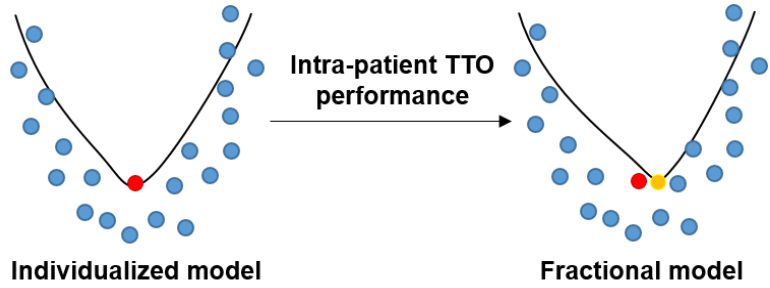

**(c) Experiment 3: intra-patient TTO application**

**Figure 3.** Experiment design. (a) Experiment 1: we compare the population model and the individualized model in terms of segmentation accuracy. (b) Experiment 2: we compare the individualized model refined from the population model with the individualized model refined from the randomized model in terms of efficiency. (c) Experiment 3: intra-patient TTO was applied to refine the first individualized model to a later fraction for performance evaluation.

In experiment 1 shown in Figure 3(a), the population model was trained on pCT and CBCT fraction 21 pairs from 200 training patients, and subsequently tested on pCT and CBCT fraction 21 pairs from the 39 test patients. Individualized models were obtained by applying inter-patient TTO to the 39 test patients, starting from the trained population model. The TTO process stops when the loss curve converges. In this and all subsequent experiments, an absolute change of less than 0.005 in the loss function is counted as no decrease, and the loss function stops decreasing in 50 iterations, defined as convergence. We compared the segmentation accuracy of the individualized TTO models and population model with different DL networks ranging from very simple to state-of-the-art neural networks including CNN, FIAM (Kuang and Schmah, 2019), Voxelmorph (Dalca et al., 2019), 5-cascaded Voxelmorph, VTN (Zhao et al., 2020), and 10-cascaded VTN (Zhao et al., 2019). CNN is a simple network that only has 10 convolutional layers without any downsampling or upsampling layers. Each convolutional layer has 16 filters and is followed by a LeakyReLU activation layer, and we added CNN architecture to our experiments to illustrate the generalizability of TTO to different types of architectures. Traditional methods including Elastix, 3DSlicer B-spline deformable registration, and 3DSlicer demon deformable registration were also performed for comparison. Since sCT and CT are considered the same image modality and used as fixed and moving images in our experiments, the loss function of all the algorithms in this study are based on intensity-based similarity metrics. We add regularization terms weighted by $\lambda$ in the loss function for stabilization purposes. Weighting factor $\lambda$ in the loss function is set to the default values used in the originally published papers, and Adam optimization with a learning rate of 0.0002 and batch size of one was used for all the TTO and population models.

Since individualized models can also be obtained by TTO starting from a randomized DL model on a new patient's data directly (one-shot learning), time efficiency was studied between inter-patient TTO starting from a population model versus starting from scratch in experiment 2, shown in Figure 3(b). The 39 test patients with pCT and CBCT fraction 21 pairs were used in this experiment. The TTO process stops when the loss curve flattens for each case during optimization. We picked the best two DL architectures for this experiment: 5-cascaded Voxelmorph and 10-casccaded VTN for time efficiency evaluation, since these two architectures are the most complicated with more time cost for each iteration during optimization, and have most accurate results among all architectures tested. These two architectures are also the only two methods that can compete with traditional DIR methods in our experiments.

In the last experiment, performance of intra-patient TTO application was studied. An individualized model was obtained by applying TTO to pCT and CBCT fraction 20 of a test patient starting from the population model. A fractional model was then obtained by applying TTO to pCT and CBCT fraction 21 of the same test patient starting from the individualized model. We repeat this process for the 39 test patients. Similar to the previous experiment, the performance gain and the optimization time for intra-patient TTO models to converge for the test patients were studied.

The best two DL architectures including 5-cascaded Voxelmorph and 10-casccaded VTN were selected for this experiment.

## 2.4. Evaluation metrics

To quantitatively evaluate segmentation accuracy, dice similarity coefficient (DSC) and 95% Hausdorff distantace (HD95) were calculated between predicted and manual segmentations. DSC is intended to gauge the similarity of the prediction and ground truth by measuring volumetric overlap between them. It is defined as

$$DSC = \frac{2|X \cap Y|}{|X|+|Y|}, \tag{17}$$

where X is the prediction, and Y is the ground truth.

HD is the maximum distance from a set to the nearest point in another set. It can be defined as

$$HD(X,Y) = \max(d_{XY}, d_{YX}) = \max\{\max_{x \in X} \min_{y \in Y} d(x,y), \max_{y \in Y} \min_{x \in X} d(x,y)\}. \tag{18}$$

HD95 is based on the 95$^{th}$ percentile of the distances between boundary points in X and Y. The purpose of this metric is to avoid the impact of a small subset of the outliers.

## 3. Results

### 3.1. Population vs. inter-patient TTO

We compared the segmentation accuracy of individualized and population models for different architectures such as a simple one like 10-layer CNN to state-of-the-art architecture like Voxelmorph and cascaded VTN. The individualized model is obtained by applying inter-patient TTO from a population model. The average dice coefficients and HD95 of the 17 selected organs and target from 39 test patients using population models and individualized models with different architectures including CNN, FAIM, Voxelmorph, VTN, 5-cascaded Voxelmorph, and 10-cascaded VTN are shown in Table 1. The performance of each of the 17 structures compared between population models and individualized models with all the tested architectures are shown in Supplementary figures 1-6. From Table 1, we can see that individualized models all have improvement from the corresponding population models for all the architecture tested in dice coefficient and HD95. The absolute improvement from the population models to the individualized models are 0.03, 0.03,0.03,0.04, 0.01, and 0.02 in dice coefficient, and 0.35 mm, 0.27 mm, 0.98 mm, 0.32 mm, 0.07 mm, and 0.06 mm in HD95 for CNN, FAIM, Voxelmorph, VTN, 5-cascaded Voxelmorph, and 10-cascaded VTN, respectively. For visual comparison, examples of auto-segmentation by a 10-cascaded population model and individualized model were plotted in Figure 4. Overall, the architectures which have poor population model performance tend to have a large performance gain with the TTO method. On the contrary, architectures with good population model performance tend to have small performance gain with the TTO method.

When comparing the performance of individualized models among all the tested architectures shown in Supplementary figure 7, the CNN individualized model has the worst performance of all, even though it has large improvement gains from the population model. While 5-cascaded Voxelmorph and 10-cascaded VTN individualized models show little improvement from population models, they still have the best performance among all the individualized models. We can see from Supplementary figure 7 that a better architecture shows enhanced performance not only in typical training strategy, but also in TTO mode. It is evident that the performance of the individualized model is architecture related, and it can be presumed that better architecture will lead to improved performance with the TTO method.

For advanced architectures like 5-cascaded Voxelmorph and 10-cascaded VTN, even though the improvement from population models by TTO are small when averaged over all the testing patients, significant performance gains can be observed for outlier patients where population models previously failed. From figure 5, we can see that there is one patient showing a 0.12 DSC gain for 5-cascaded Voxelmorph and a 0.11 DSC gain for 10-cascaded VTN by the TTO method. The number of patients who have at least a 0.05 DSC improvement or 2 mm HD95 improvement by inter-patient TTO for CNN, FAIM, Voxelmorph, VTN, 5-cascaded Voxelmorph, and 10-cascaded VTN architectures are 5, 6, 10, 9, 2 and 2 out of 39 test patients, respectively. Thus, models generated by the TTO method are less vulnerable to generalizability problems.

We also compared individualized models to traditional DIR methods including 3DSlicer B-spline deformable registration, 3DSlicer Demon deformable registration, and Elastix, shown in Table 2. We only picked the best two architectures for comparison: 5-cascaded Voxelmorph and 10-casscaded VTN, since only these two architectures can compete with traditional methods. The 5-cascaded Voxelmorph and 10-cascaded VTN individualized models have higher average dice coefficients and lower HD95 values than traditional DIR methods over all 17 structures from 39 test patients. Since Elastix has the best performance among traditional methods, only Elastix is plotted in Supplementary figure 7 for comparison with individualized models in each structure. The auto-segmentation generated by Elastix and 10-cascaded VTN individualized models from some test patients are shown in Figure 4 for visual evaluation. It is noticeable that the discrepancy between contours generated by individualized models and ground truth contours are much less than those between contours generated by Elastix and ground truth. Overall, 5-cascaded Voxelmorph and 10-cascaded VTN using our proposed TTO method can outperform traditional methods.

**Table 1. The average DSC and HD95 of the 13 selected structures from the 39 test patients for different DL architectures with population models and individualized models.** Individualized models are obtained by applying inter-patient TTO from a population model. The green numbers are the absolute improvements from the population model to the individualized model.

| DL Architecture | DSC | | | HD95 (mm) | | |
| --- | --- | --- | --- | --- | --- | --- |
| | Population model | Individualized model | Improvement | Population model | Individualized model | Improvement |
| CNN | 0.78 | 0.81 | +0.03 | 3.18 | 2.83 | -0.35 |
| FAIM | 0.80 | 0.83 | +0.03 | 2.88 | 2.61 | -0.27 |
| Voxelmorph | 0.79 | 0.82 | +0.03 | 3.87 | 2.89 | -0.98 |
| VTN | 0.81 | 0.85 | +0.04 | 2.82 | 2.50 | -0.32 |
| 5-cascaded Voxelmorph | 0.83 | 0.84 | +0.01 | 2.53 | 2.46 | -0.07 |
| 10-cascaded VTN | 0.83 | 0.85 | +0.02 | 2.39 | 2.33 | -0.06 |

**Table 2. Comparison between traditional DIR methods and individualized models with state-of-the-art DL architectures.** Numbers in this table were calculated by the average DSC and HD95 of the 13 selected structures from the 39 test patients.

| | 3DSlicer Demon | 3DSlicer B-spline | Elastix | 5-cascaded Voxelmorph individualized model | 10-cascaded VTN individualized model |
| --- | --- | --- | --- | --- | --- |
| DSC | 0.78 | 0.82 | 0.83 | **0.84** | **0.85** |
| HD95 (mm) | 3.55 | 2.82 | 2.83 | **2.46** | **2.33** |

| CBCT with ground truth | Elastix | 10-cascaded VTN population model | 10-cascaded VTN individualized model |
| --- | --- | --- | --- |

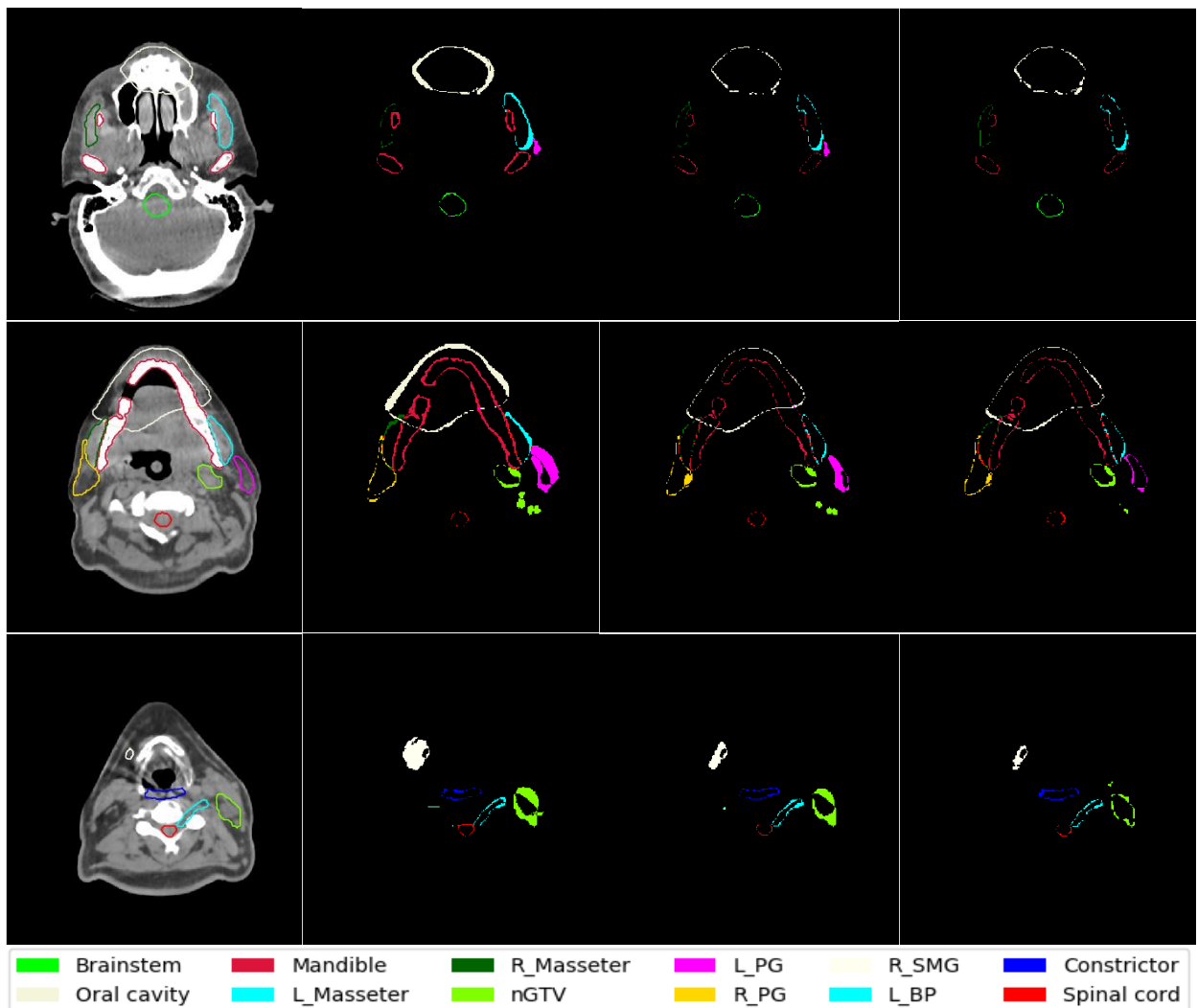

**Figure 4. Contours of test patients from axial view.** The images from left to right are CBCT with manual ground truth contours on it, the discrepancy between Elastix contours and ground truth contours, the discrepancy between 10-cascaded VTN population model contours and ground truth contours, and the discrepancy between 10-cascaded VTN individualized model contours and ground truth contours respectively.

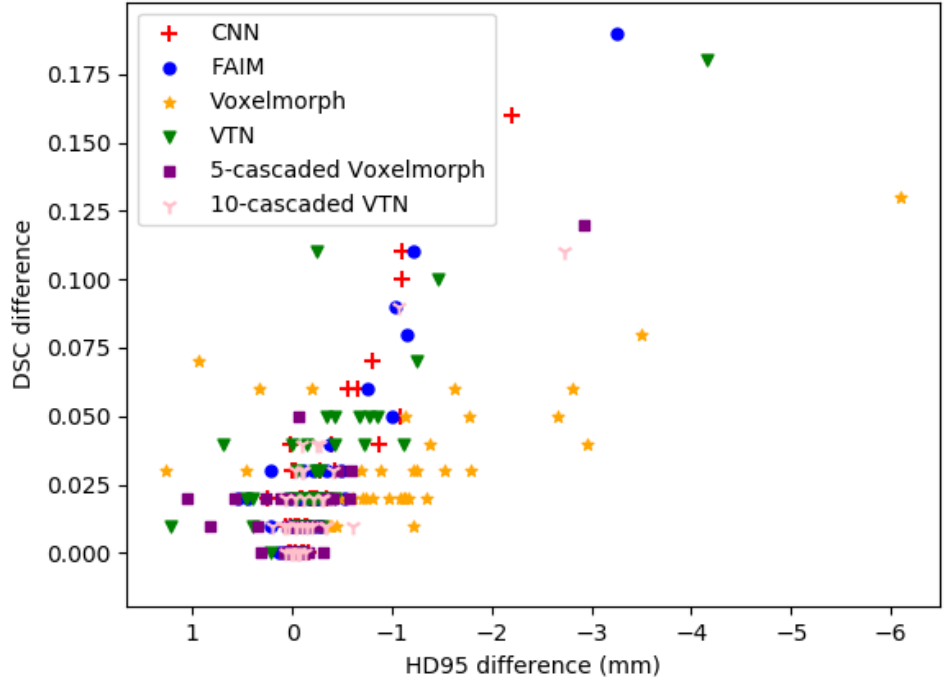

**Figure 5. Distribution of DSC and HD95 change from a population model to an individualized model for 39 test patient.** X axis and Y axis are the average HD95 difference and average DSC difference of 17 structures between a population model and an individualized model for a patient.

## 3.2. Inter-patient TTO: starting from a population model vs. starting from a randomized model

In inter-patient TTO application, we refine a population model to a new patient instead of starting TTO from scratch for efficiency gain. The time cost for inter-patient TTO starting from a randomized model and a population model is plotted in Figure 6. The average time cost for inter-patient TTO starting from a randomized model is 10.25 hours for 5-cascaded Voxelmorph and 9.44 hours for 10-cascaded VTN. However, the convergence time is dramatically decreased to 3.78 minutes and 3.60 minutes, respectively, by starting from a population model. The majority of test patients can have a personalized model ready in four minutes, and the maximum time cost is no more than 10 minutes. Figure 7 shows an example of how the auto-segmentation improves with time during inter-patient TTO. Sharp improvement happened at approximately three minutes, and afterwards, the improvement starts to slow down and finally becomes unnoticeable. Therefore, the time needed by inter-patient TTO for model performance gain is clinically acceptable.

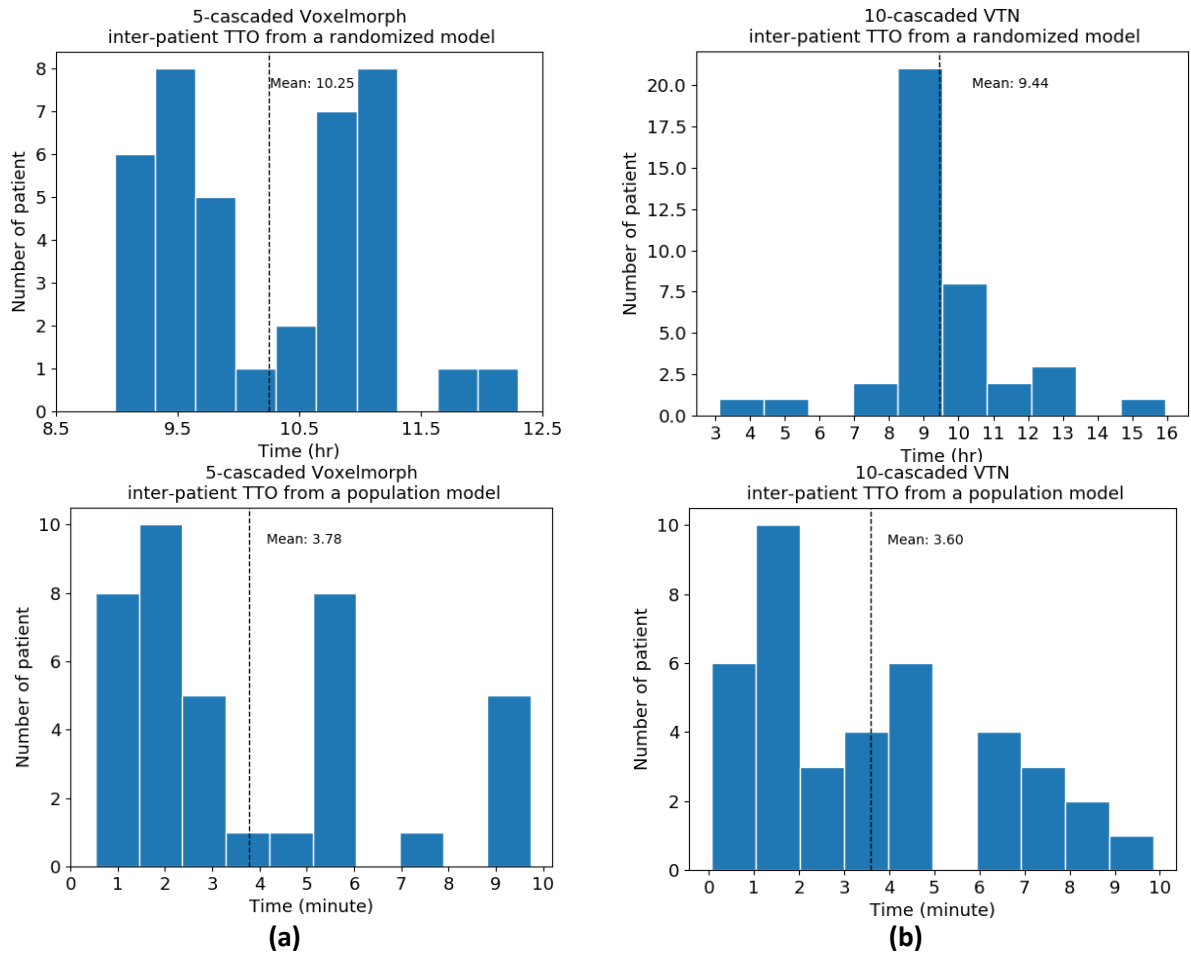

**Figure 6. The time cost for inter-patient TTO.** Two architectures were tested: (a) 5-cascaded Voxelmorph and (b) 10-cascaded VTN.

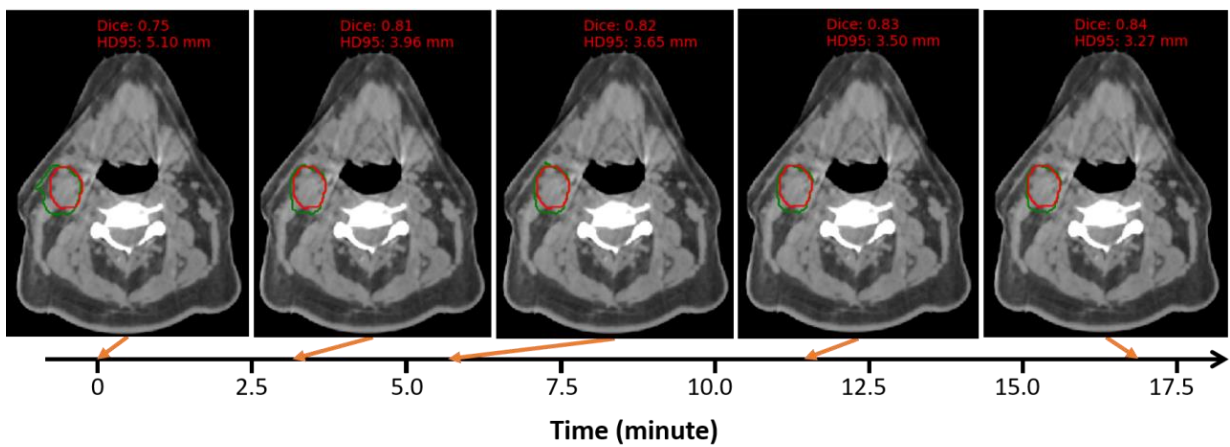

**Figure 7. One example of auto-segmentation performance is improved as time goes by with inter-patient TTO starting from a population moel.** The architecture used here is 10-cascaded VTN. Background image is CBCT and the structure contoured is nGTV. Red is the ground truth contour and green is the TTO contour.

## 3.3 Intra-patient TTO

In intra-patient TTO application, the time cost for the intra-patient TTO model to converge starting from an individualized model is plotted in Figure 8. The average intra-patient TTO time for 39 test patients is 1.06 minutes with 5-cascaded Voxelmorph architecture and 1.24 minutes with 10-cascaded VTN architecture. The majority of test patients can have a personalized model ready in one minute, and the maximum time cost is no more than six minutes. The segmentation performance gain through intra-patient TTO is not obvious, shown in Table 3, and Figure 9 shows an example of how the auto-segmentation improves with time during intra-patient TTO. Sharp improvement happened at approximately 1.5 minutes, where posterior node appears improved and inferior node is visually changed little. While after that, the improvement is negligible. The individualized model that has been optimized on a pair of images from only one fraction from a specific patient can work well on another pair of images from the next fraction. This means very few or no iterations may be needed between fractions within a patient.

**Table 3. Average DSC of 39 test patients with 13 selected structures before and after intra-patient TTO**

|  | Before intra-patient TTO | After intra-patient TTO |
|---|---|---|
| **5-cascaded Voxelmorph** | 0.839 | 0.842 |
| **10-cascaded VTN** | 0.841 | 0.842 |

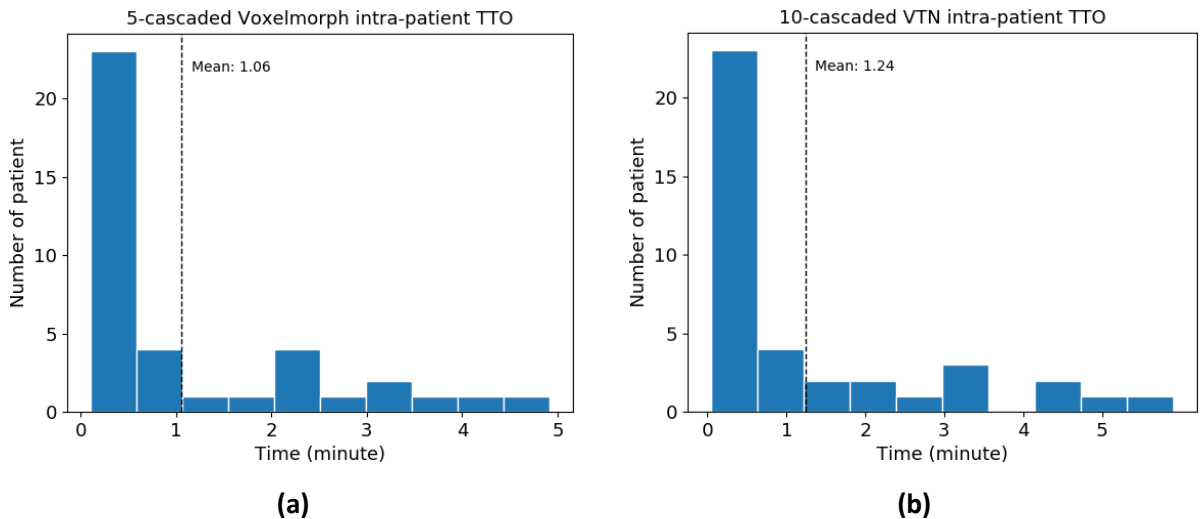

**Figure 8. The time cost for intra-patient TTO starting from an individualized model.** Two architectures were tested: (a) 5-cascaded Voxelmorph and (b) 10-cascaded VTN.

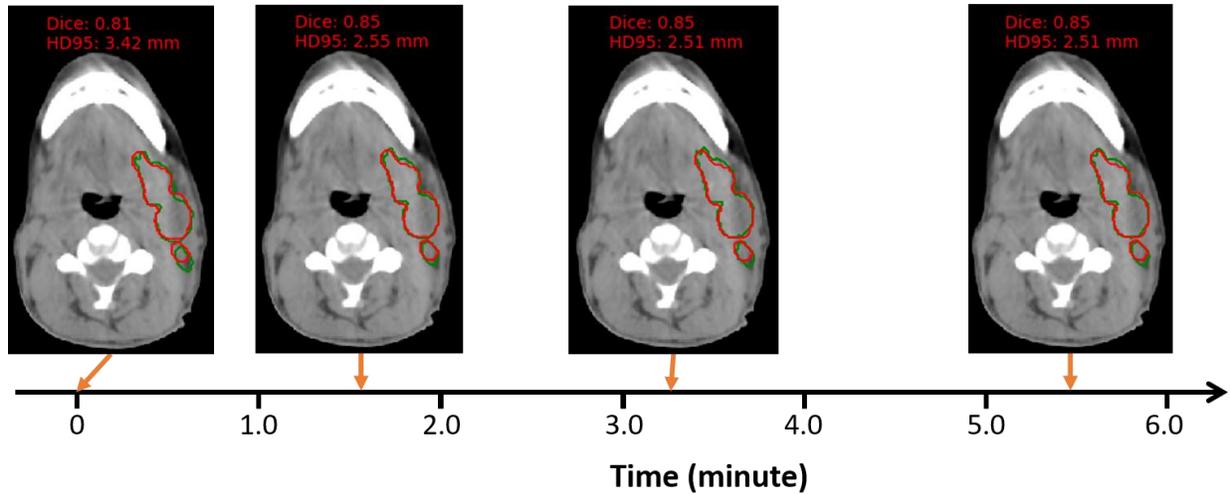

**Figure 9. One example of auto-segmentation performance change with intra-patient TTO.** The architecture used here is 10-cascaded VTN. Background image is CBCT and the structure contoured is nGTV. Red is the ground truth contour and green is the TTO contour.

## 4. Discussion and Conclusion

In this experiment, we proposed a new concept called TTO that attempts to improve online ART auto-segmentation by individualizing baseline population models for each patient and also based on the changing fractional anatomy at each CBCT acquired for that patient during treatment. We further demonstrated that this TTO method improved the performance of the population models for all architectures tested. In general, better baseline DL architectures using the TTO method outperformed the traditional DIR methods in terms of accuracy and robustness. Given one of the main burdens of online ART is the time required to review and edit deformed or auto-segmented contours by the provider, we find these initial results showing the added benefit of refining and individualizing baseline population models by incorporating data from the changing fractional anatomy of independent patients to be promising.

Although one concern about re-training models at each fraction would be the time required for this process that may make this impractical, we found this TTO method to be efficient. In inter-patient applications, it took about 3 minutes for a model to converge starting from a pre-trained population model in the inter-patient application. In intra-patient applications, it took about 1 minute for a model to converge starting from a pre-trained individualized model. Compared to one shot DIR learning, which starts to train a model from scratch using only a pair of moving and fixed image, TTO can dramatically decrease the time cost and make it feasible for online applications such as CBCT-based online ART.

One advantage of the TTO method is its flexibility, as it can be applied to any unsupervised DIR neural network. The TTO optimization process is also very easy since the hyper-parameters used in the deep neural networks can be fixed for all the cases, avoiding the parameter tuning process in traditional DIR methods.

Another main advantage of the TTO method is its ability to improve model generalizability. A population model can be adapted to each individual patient by TTO rather than a same population model applied to all patients. Each individualized model can further be adapted from fraction to fraction through the ART course. In the case when the population model fails, TTO-adapted models can boost model performance significantly. We showed a patient case in Supplementary Figure 8(d) and (e) in which the population model totally failed in delineating the constrictor and larynx at the border slices, whereas the TTO model can prove successful in these instances. In the other figures in

Supplementary Figure 8, it is clear that the TTO model significantly improves segmentation accuracy compared to the population model.

All experiments in this work were performed by one NVIDIA V100 GPU with 32 GB RAM. The time cost per iteration during TTO was around three seconds for 5-cascaded Voxelmorph and 10-cascaded VTN architecture. Since the number of iterations needed for a population model to converge on a specific patient's data to generate an individualized model is so low, inter-patient TTO can be completed online. Furthermore, the anatomy difference between two consecutive fractions for the same patient is very small, allowing for intra-patient TTO to be accomplished in real time.

In conclusion, we designed a novel TTO strategy to achieve patient and treatment fraction-specific models for image registration and structure propagation to facilitate CBCT-based online ART workflow.

## Code availability

The code for the deep learning frameworks, and the software packages used in the experiments are all publicly available. FAIM is available at https://github.com/dykuang/Medical-image-registration , Voxelmorph is available at https://github.com/voxelmorph/voxelmorph ,VTN and 10-cascaded VTN are available at https://github.com/microsoft/Recursive-Cascaded-Networks. The Elastix package can be downloaded from https://elastix.lumc.nl/ , and 3DSlicer software can be downloaded from https://www.slicer.org/.

## Data availability

All datasets were collected from one institution and are non-public. In agreement with HIPAA, access to the dataset will be granted on a case-by-case basis in response to requests sent to the corresponding authors and the institution.

## Acknowledgement

We would like to thank the Varian Medical Systems Inc. for supporting this study and Ms. Sepeadeh Radpour for editing the manuscript.

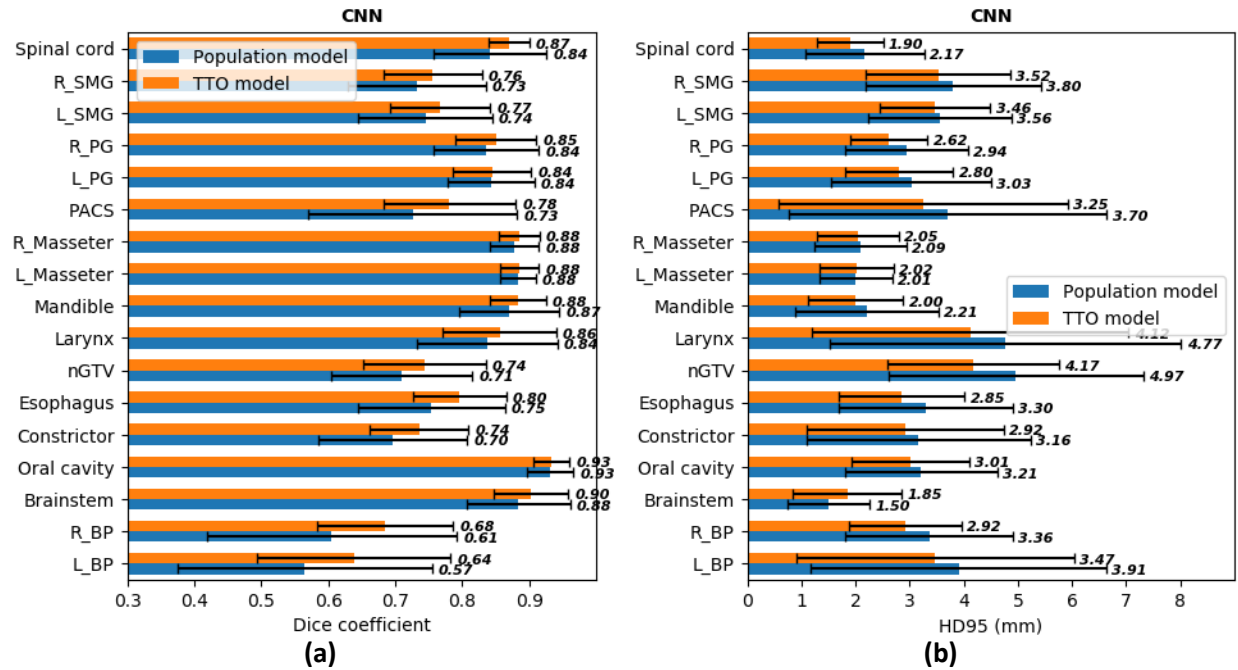

**Supplementary figure 1. Population model vs. individualized model for CNN architecture.** DSC and HD95 were calculated against the ground truth for each of the 17 selected structures from the 39 test patients. The bar plots represent mean values and the error bars represent standard deviation.

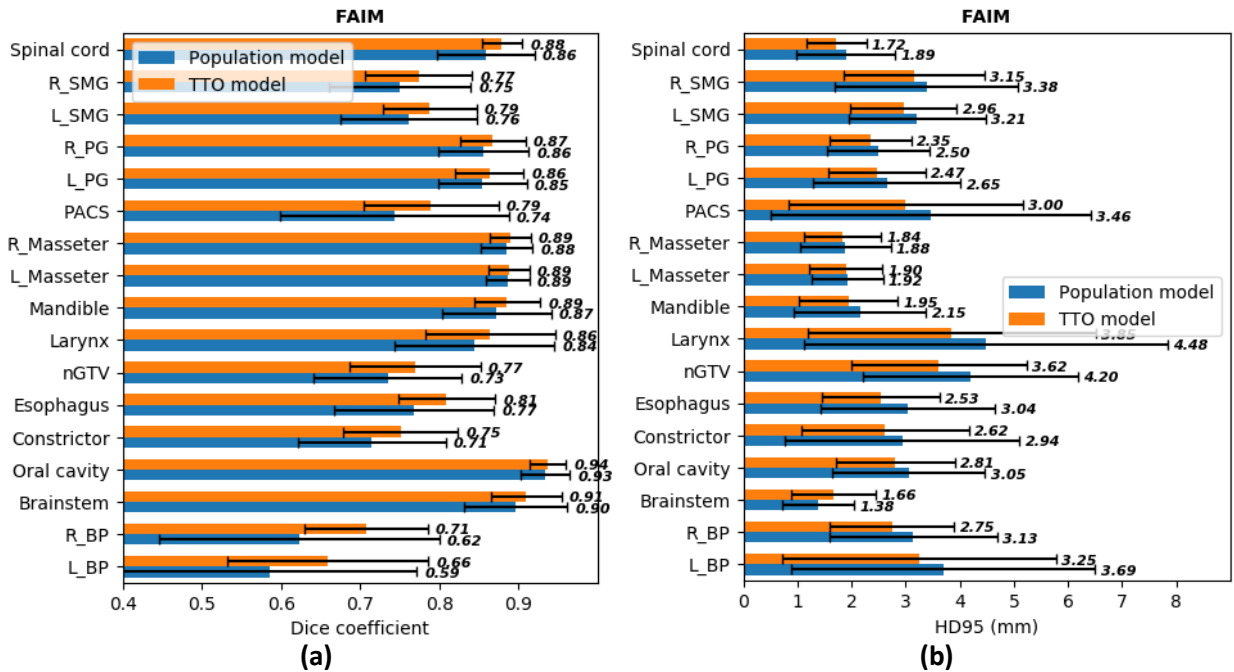

**Supplementary figure 2. Population model vs. individualized model for FAIM architecture.** DSC and HD95 were calculated against the ground truth for each of the 17 selected structures from the 39 test patients. The bar plots represent mean values and the error bars represent standard deviation.

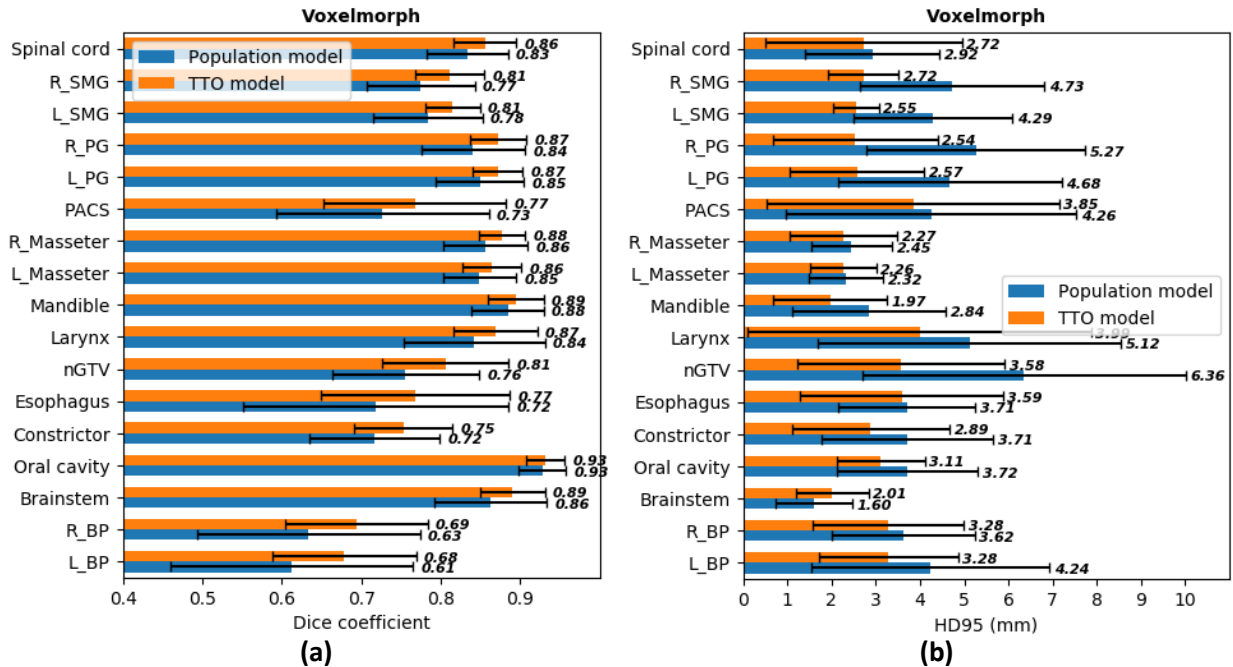

**Supplementary figure 3. Population model vs. individualized model for Voxelmorph architecture.** DSC and HD95 were calculated against the ground truth for each of the 17 selected structures from the 39 test patients. The bar plots represent mean values and the error bars represent standard deviation.

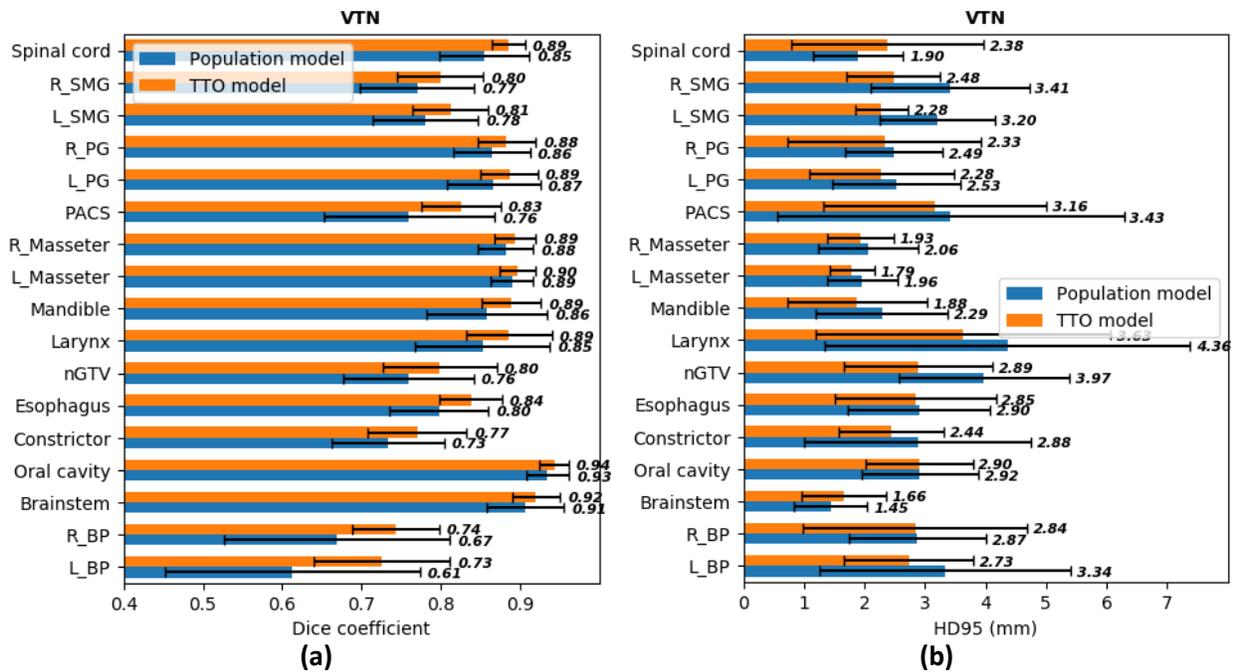

**Supplementary figure 4. Population model vs. individualized model for VTN architecture.** DSC and HD95 were calculated against the ground truth for each of the 17 selected structures from the 39 test patients. The bar plots represent mean values and the error bars represent standard deviation.

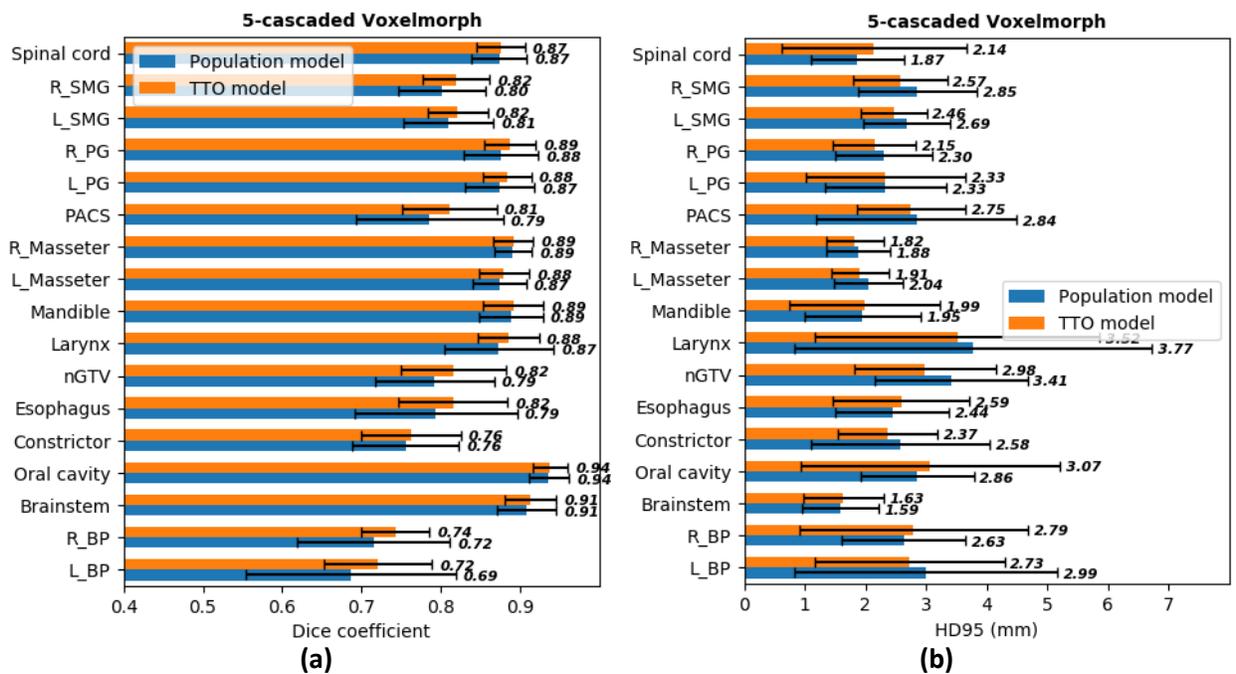

**Supplementary figure 5. Population model vs. individualized model for 5-cascaded Voxelmorph architecture.** DSC and HD95 were calculated against the ground truth for each of the 17 selected

structures from the 39 test patients. The bar plots represent mean values and the error bars represent standard deviation.

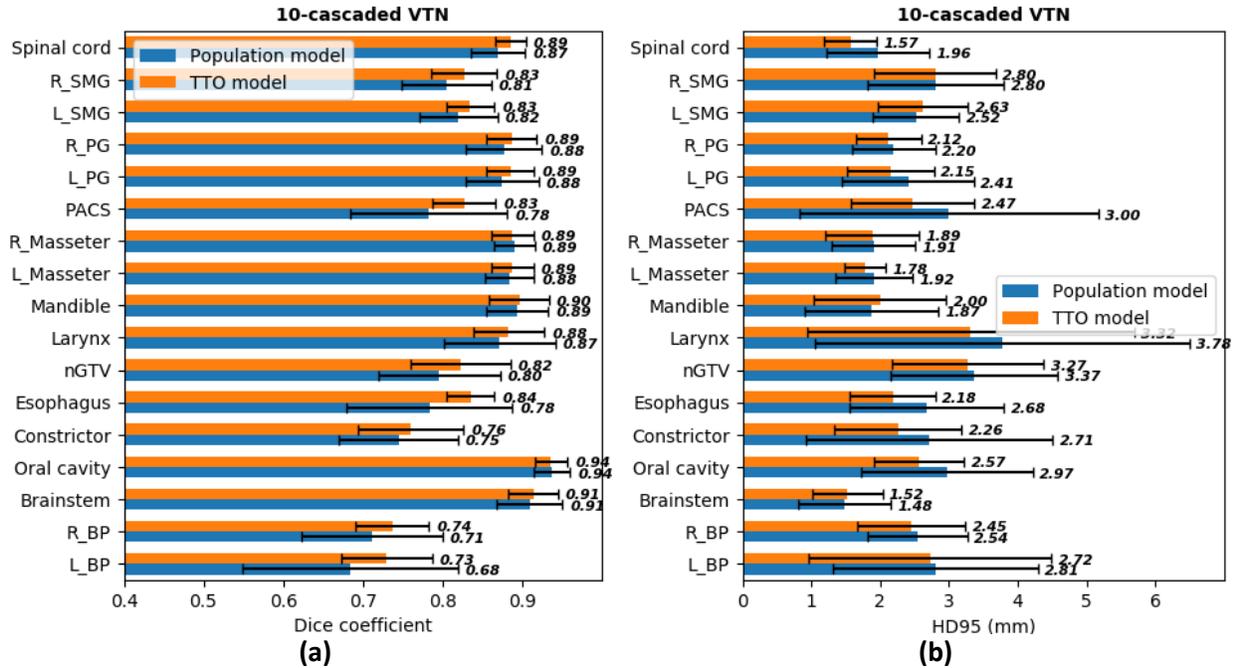

**Supplementary figure 6. Population model vs. individualized model for 10-cascaded VTN architecture.** DSC and HD95 were calculated against the ground truth for each of the 17 selected structures from the 39 test patients. The bar plots represent mean values and the error bars represent standard deviation.

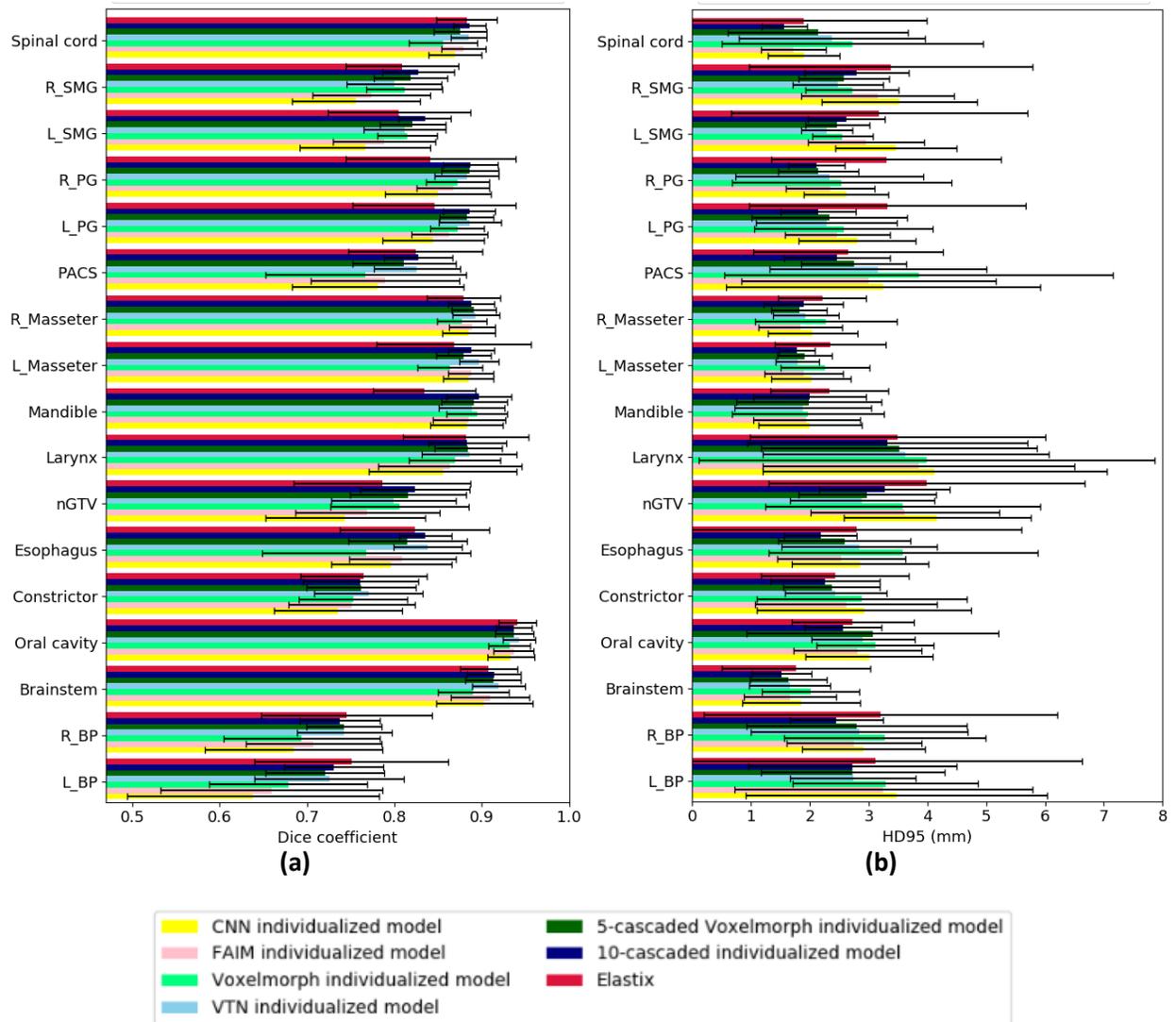

**Supplementary figure 7. Quantitative performance comparison among individualized models with different architectures and traditional DIR method.** The different architectures are CNN, FAIM, Voxelmorph, VTN, 5-cascaded Voxelmorph, and 10-cascaded VTN. DSC and HD95 were calculated against the ground truth for each of the 17 selected structures from the 39 test patients. The bar plots represent mean values and the error bars represent standard deviation.

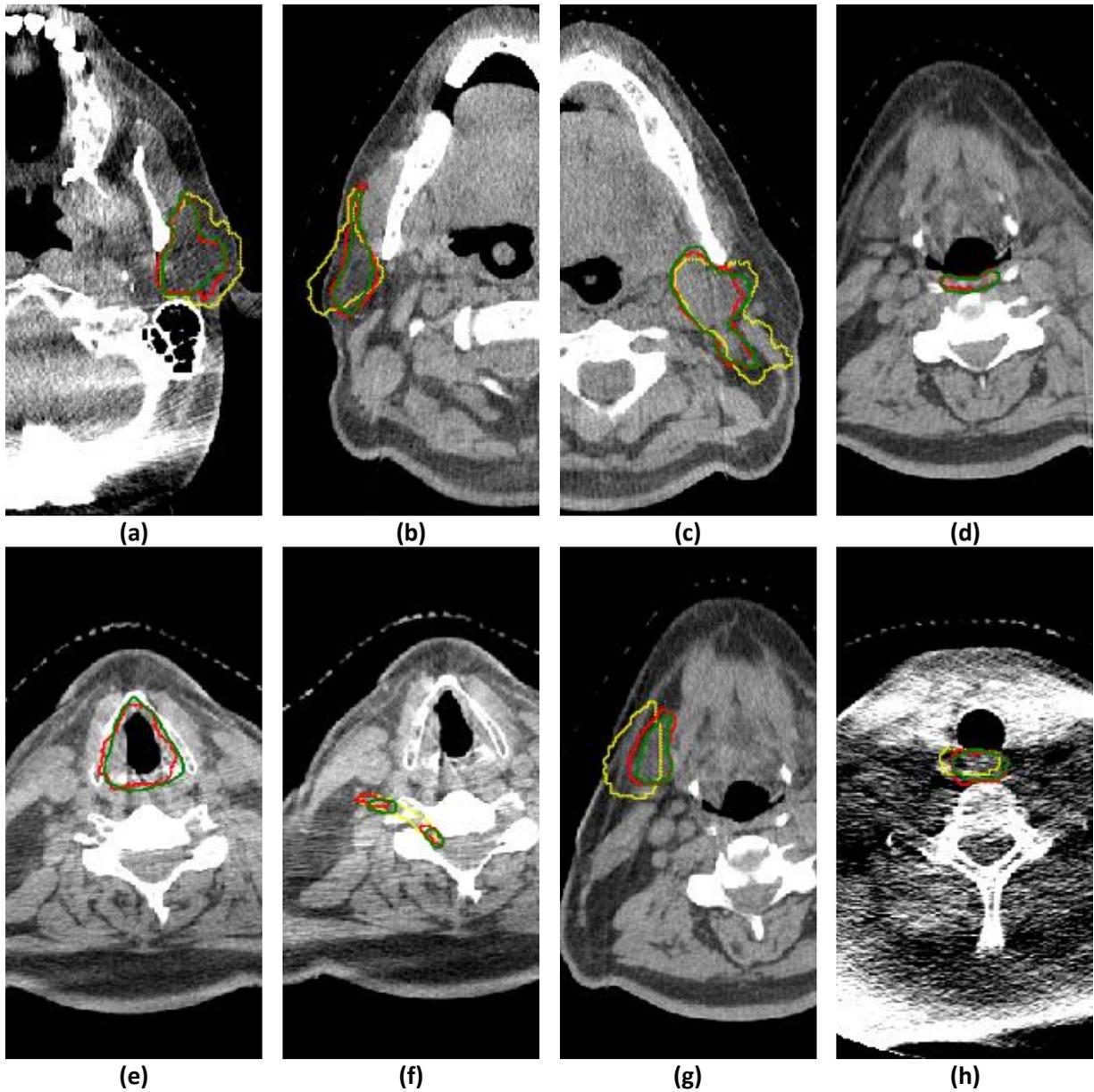

**Supplementary figure 8. Illustration of the generalization problem a population model has, showing a TTO model can avoid this problem.** The segmentations from (a) to (h) are: (a) L_PG, (b) R_PG, (c) nGTV, (d) Constrictor, (e) Larynx, (f) R_BP, (g) R_SMR, and (h) Esophagus. Green, red, and yellow contours represent ground truth, TTO model segmentation and population model segmentation, respectively. The architecture used in this experiment is Voxelmorph.